\documentclass[10pt,twocolumn, letterpaper]{article} % use this option for final submission!

\newcommand{\forReview}{0} % set to zero if it is not review mode

\makeatother
\usepackage{color}
\usepackage{ifthen}
\usepackage{float}
\usepackage{alltt}
\usepackage{newlfont} % for Box
\usepackage{wrapfig}
\usepackage{booktabs}
\usepackage{multirow}
\usepackage{comment}
\usepackage{gensymb}
\usepackage{pifont}
\usepackage{xcolor}

\usepackage{comment}
\usepackage{amsmath}
\usepackage{amssymb}
\usepackage{amsthm}

\usepackage{cvpr}
\usepackage{times}
\usepackage{epsfig}
\usepackage{graphicx}
\usepackage{subcaption}
\usepackage{microtype}
\usepackage{blindtext}

\floatstyle{plain}

\ifthenelse{\equal{\forReview}{1}}
{
\usepackage[pagebackref=true,breaklinks=true,letterpaper=true,colorlinks,bookmarks=false]{hyperref}
\ifcvprfinal\pagestyle{empty}\fi
}
{
\usepackage[breaklinks=true,colorlinks,bookmarks=false]{hyperref}
\cvprfinalcopy % *** Uncomment this line for the final submission
\ifcvprfinal\pagestyle{empty}\fi

\pagecolor{white}

\author{Shunsuke Saito\textsuperscript{1,3}
\hspace{0.3in} Tomas Simon\textsuperscript{2}
\hspace{0.3in} Jason Saragih\textsuperscript{2}
\hspace{0.3in} Hanbyul Joo\textsuperscript{3}
\vspace{5pt}
\\
\textsuperscript{1}{University of Southern California}
\hspace{0.3in} \textsuperscript{2}{Facebook Reality Labs}
\hspace{0.3in} \textsuperscript{3}{Facebook AI Research} \\ 
}%author
}

\ifcvprfinal\fi
\def\cvprPaperID{5870}

\graphicspath{
{figures}
{images}
}

\begin{document}

%%%%%%%%% TITLE
\title{PIFuHD: Multi-Level Pixel-Aligned Implicit Function for\\ High-Resolution 3D Human Digitization\thanks{Website: \url{https://shunsukesaito.github.io/PIFuHD/}}}

\maketitle
\thispagestyle{empty}

%%%%%%%%% ABSTRACT
% !TEX root = draft.tex

\begin{abstract}{~}
Recent advances in image-based 3D human shape estimation have been driven by the significant improvement in representation power afforded by deep neural networks. Although current approaches have demonstrated the potential in real world settings, they still fail to produce reconstructions with the level of detail often present in the input images. We argue that this limitation stems primarily form two conflicting requirements; accurate predictions require large context, but precise predictions require high resolution. Due to memory limitations in current hardware, previous approaches tend to take low resolution images as input to cover large spatial context, and produce less precise (or low resolution) 3D estimates as a result. We address this limitation by formulating a multi-level architecture that is end-to-end trainable. A coarse level observes the whole image at lower resolution and focuses on holistic reasoning. This provides context to an fine level which estimates highly detailed geometry by observing higher-resolution images. We demonstrate that our approach significantly outperforms existing state-of-the-art techniques on single image human shape reconstruction by fully leveraging 1k-resolution input images. 
\end{abstract}

\section{Introduction}
\label{sec:intro}

High-fidelity human digitization is the key to enabling a myriad of applications from medical imaging to virtual reality. While metrically accurate and precise reconstructions of humans is now possible with multi-view systems~\cite{guo2019relightables, lombardi2018deep}, it has remained largely inaccessible to the general community due to its reliance on professional capture systems with strict environmental constraints (e.g., high number of cameras, controlled illuminations) that are prohibitively expensive and cumbersome to deploy. Increasingly, the community has turned to using high capacity deep learning models that have shown great promise in acquiring reconstructions from even a single image~\cite{hmrKanazawa17, varol18_bodynet, natsume2019siclope, alldieck2019learning}. However, the performance of these methods currently remains significantly lower than what is achievable with professional capture systems. 

\begin{figure}[t]
\begin{center}
  \includegraphics[width=\linewidth]{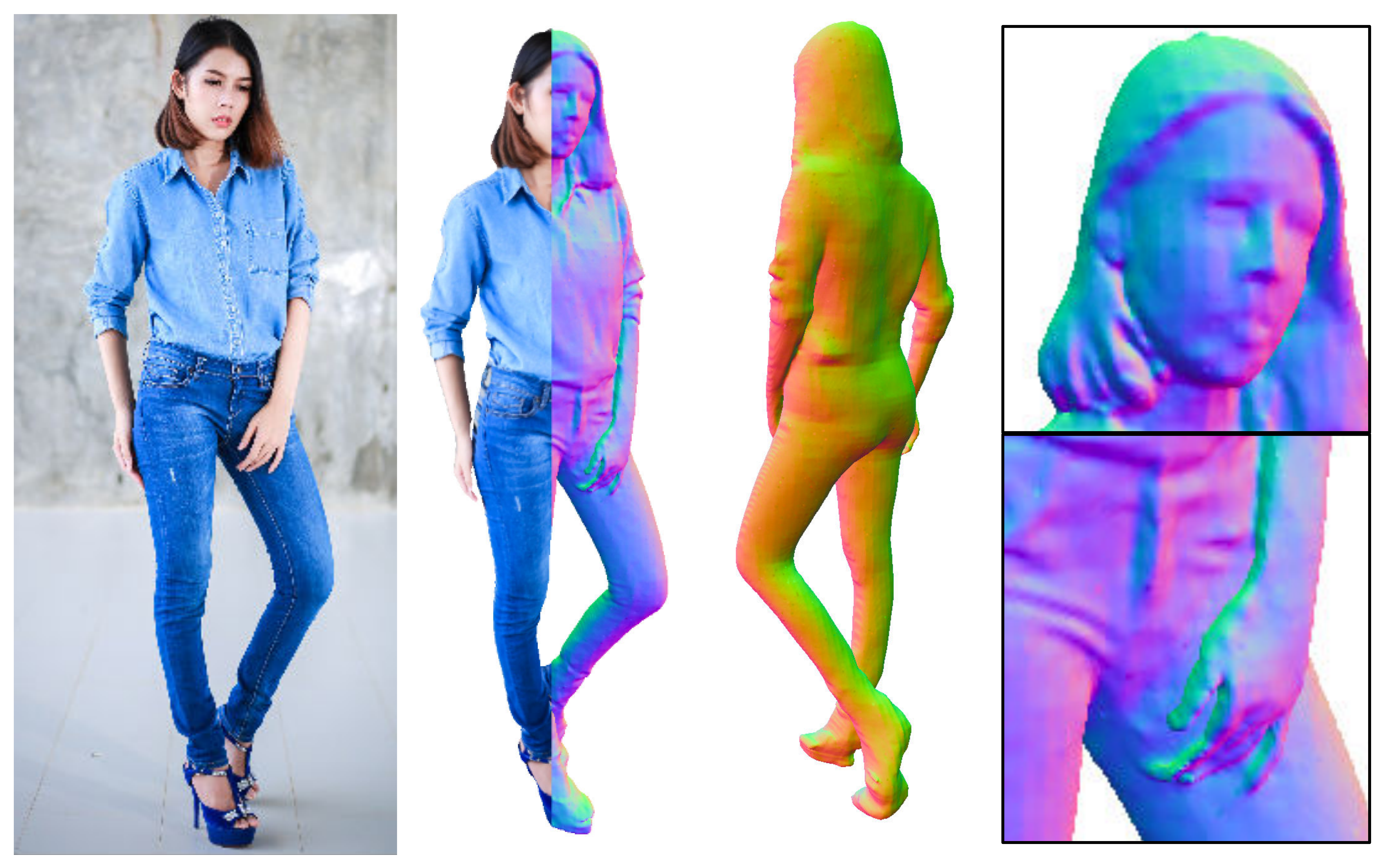}
\end{center}
\vspace{-10pt} 
\caption{Given a high-resolution single image of a person, we recover highly detailed 3D reconstructions of clothed humans at 1k resolution.}
\label{fig:teaser}
\end{figure}

The goal of this work is to achieve high-fidelity 3d reconstruction of clothed humans from a single image at a resolution sufficient to recover detailed information such as fingers, facial features and clothing folds (see Fig.~\ref{fig:teaser}). Our observation is that existing approaches do not make full use of the high resolution (e.g., 1k or larger) imagery of humans that is now easily acquired using commodity sensors on mobile phones.
This is because the previous approaches rely on holistic reasoning to map between the 2D appearance of an imaged human and their 3D shape, where, in practice, down-sampled images are used due to the prohibitive memory requirements~\cite{hmrKanazawa17, varol18_bodynet}. Although local image patches have important cues for detailed 3D reconstruction, these are rarely leveraged in the full high-resolution inputs due to the memory limitations of current graphics hardware.

Approaches that aim to address this limitation can be categorized into one of two camps. In the first camp, the problem is decomposed in a coarse-to-fine manner, where high-frequency details are embossed on top of low-fidelity surfaces. In this approach, a low image resolution is used to obtain a coarse shape. Then, fine details represented as surface normal \cite{Zheng_2019_ICCV} or displacements \cite{Alldieck_2019_ICCV} are added by either a post-process such as Shape From Shading \cite{horn1970shape} or composition within neural networks. The second camp employs high-fidelity models of humans (e.g., SCAPE~\cite{anguelov2005scape}) to hallucinate plausible detail. Although both approaches result in reconstructions that appear detailed, they often do not faithfully reproduce the true detail present in the input images. 

In this work, we introduce an end-to-end multi-level framework that infers 3D geometry of clothed humans at an unprecedentedly high 1k image resolution in a pixel-aligned manner, retaining the details in the original inputs without any post-processing. Our method differs from the coarse-to-fine approaches in that no explicit geometric representation is enforced in the coarse levels. Instead, implicitly encoded geometrical context is propagated to higher levels without making an explicit determination about geometry prematurely. We base our method on the recently introduced Pixel-Aligned Implicit Function (PIFu) representation~\cite{saito2019pifu}. The pixel-aligned nature of the representation allows us to seamlessly fuse the learned holistic embedding from coarse reasoning with image features learned from the high-resolution input in a principled manner. Each level incrementally incorporates additional information missing in the coarse levels, with the final determination of geometry made only in the highest level.

Finally, for a complete reconstruction, the system needs to recover the backside, which is unobserved in any single image. As with low resolution input, missing information that is not predictable from observable measurements will result in overly smooth and blurred estimates. We overcome this problem by leveraging image-to-image translation networks to produce backside normals, similar to \cite{natsume2019siclope, gabeur:hal-02242795, Smith_2019_ICCV}. Conditioning our multi-level pixel-aligned shape inference with the inferred back-side surface normal removes ambiguity and significantly improves the perceptual quality of our reconstructions with a more consistent level of detail between the visible and occluded parts. 

The main contributions in this work consists of:
\begin{itemize}
    \item an end-to-end trainable coarse-to-fine framework for implicit surface learning for high-resolution 3D clothed human reconstruction at 1k image resolution.
    \item a method to effectively handle uncertainty in unobserved regions such as the back, resulting in complete reconstructions with high detail.
\end{itemize}
% !TEX root = draft.tex

\section{Related Work}
\label{sec:related_work}

\paragraph{Single-View 3D Human Digitization}
Single-view 3D human reconstruction is an ill-posed problem due to the fundamental depth ambiguity along camera rays. To overcome such ambiguity, parametric 3D models~\cite{anguelov2005scape, loper2015smpl, joo2018total, pavlakos2019expressive} are often used to restrict estimation to a small set of model parameters, constraining the solution space to a specifically chosen parametric body model~\cite{bogo2016keep, lassner2017unite, kanazawa2018end, xiang2019monocular, pavlakos2019expressive, xu2019denserac}. However, the expressiveness of the resulting models is limited by using a single template mesh as well as by the data on which the model is built (often comprised mainly of minimally clothed people). While using a separate parameteric model can alleviate the limited shape variation~\cite{bhatnagar2019multi}, large deformations and topological changes are still non-trivial to handle with these shape representations.

Researchers have also proposed methods that do not use parametric models, but rather directly regress ``free-form'' 3D human geometry from single views. These approaches vary their directions based on the input and output representation that each algorithm uses. Some methods represent the 3D output world via a volumetric representation~\cite{varol18_bodynet}. Of particular relevance to this work is the DeepHuman~\cite{Zeng_2019_ICCV} approach of Zheng et al., where a discretized volumetric representation is produced by the network in increasing resolution and detail. Additional details using surface normals are embossed at the final level. While this method obtains impressive results, the cubic memory requirement imposed by the discrete voxel representation prevents obtaining high resolution simply by naively scaling the input resolution. Alternative methods 
consider additional free-form deformation on top of a parametric model space~\cite{alldieck2019learning}, and there exist also multiple approaches that predict depth maps of the target people as output~\cite{tang2019neural, gabeur:hal-02242795, Smith_2019_ICCV}. 

The recently introduced Pixel-Aligned Implicit Function (PIFu)~\cite{saito2019pifu} does not explicitly discretize the output space representation but instead regresses a function which determines the occupancy for any given 3D location. This approach shows its strength in reconstructing high-fidelity 3D geometry without having to keep a discretized representation of the entire output volume in memory simultaneously. Furthermore, unlike implicit surface representations using a global feature vector \cite{mescheder2018occupancy, park2019deepsdf, chen2018implicit_decoder}, PIFu utilizes fully convolutional image features, retaining local details present in an input image. 

\paragraph{High-Resolution Synthesis in Texture Space}
A number of recent approaches pursue reconstructing high-quality 3D texture or geometry by making use of a texture map representation~\cite{yamaguchi2018high, tran2018extreme, lazova3dv2019} on which to estimate geometric or color details. Particularly, the Tex2Shape approach of Alldieck et al.~\cite{Alldieck_2019_ICCV} aims to reconstruct high quality 3D geometry by regressing displacements in an unwrapped UV space. However, this type of approach is ultimately limited by the topology of the template mesh (exhibiting problems when representing different topologies, such as required by different hair styles or skirts) and the topology chosen for the UV parameterization (e.g., visible seam artifacts around texture seams).  Recent approaches leverage neural network models to predict intermediate texture or depth representations that are then used to reconstruct final 3D geometry output~\cite{sela2017unrestricted, Zeng_2019_ICCV}.

Our work is also related to approaches that produce high quality or high resolution synthetic human images. Recent methods consider producing high quality synthetic faces to overcome limitations of original GAN-based approaches~\cite{wang2018high, karras2017progressive}. Similar trade-offs are pursued in semantic segmentation tasks~\cite{chen2017deeplab, deeplabv3plus2018}. 

\section{Method}
\label{sec:method}

\begin{figure*}[t]
\centering{
\includegraphics[width=\linewidth]{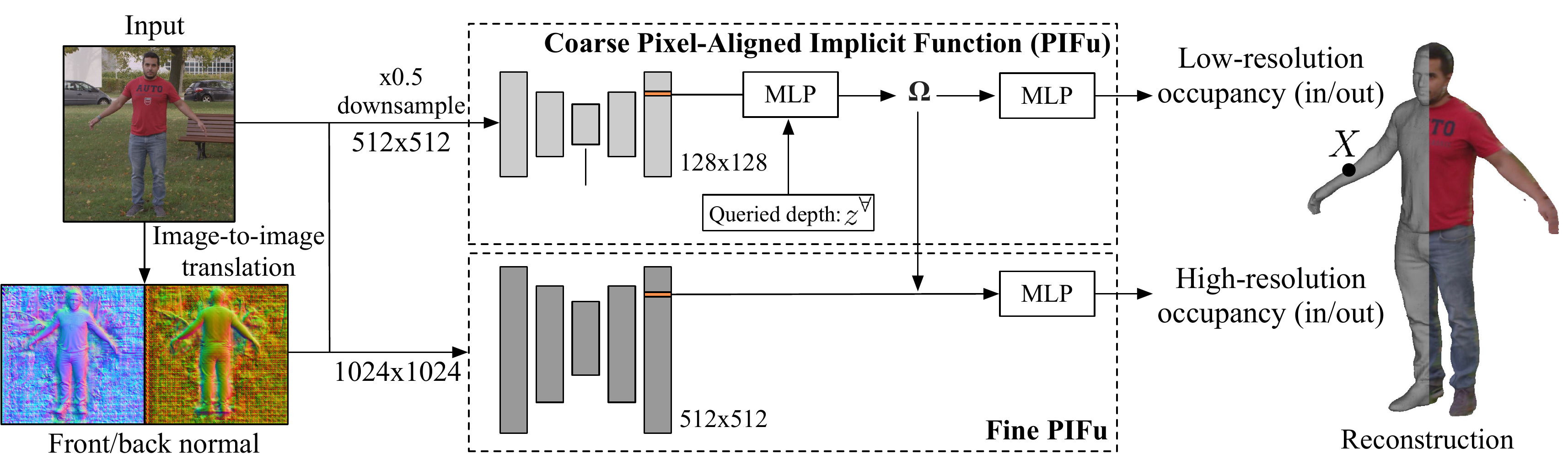}}
\vspace{-7pt}
\caption{Overview of our framework. Two levels of pixel-aligned predictors produce high-resolution 3D reconstructions. The coarse level (top) captures global 3D structure, while high-resolution detail is added by the fine level.}
\label{fig:overview}
\vspace{-10pt}
\end{figure*}

Our method builds on the recently introduced Pixel-aligned Implicit Function (PIFu) framework of~\cite{saito2019pifu}, which takes images with resolution of 512$\times$512 as input and obtains low-resolution feature embeddings (128$\times$128). To achieve higher resolution outputs, we stack an additional pixel-aligned prediction module on top of this framework, where the fine module takes as input higher resolution images (1024$\times$1024) and encodes into high-resolution image features (512$\times$512). The second module takes the high-resolution feature embedding as well as the 3D embeddings from the first module to predict an occupancy probability field. To further improve the quality and fidelity of the reconstruction, we first predict normal maps for the front and back sides in image space, and feed these to the network as additional input. See Fig.~\ref{fig:overview} for an overview of the method.

\subsection{Pixel-Aligned Implicit Function}
\label{sec:pifu}
We briefly describe the fundation of PIFu introduced in \cite{saito2019pifu}, which constitutes the coarse level of our method (upper half in Fig.~\ref{fig:overview}). The goal of 3D human digitization can be achieved by estimating the occupancy of a dense 3D volume, which determines whether a point in 3D space is inside the human body or not. In contrast to previous approaches, where the target 3D space is discretized and algorithms focus on estimating the occupancy of each voxel explicitly (e.g.,~\cite{Zheng_2019_ICCV}), the goal of PIFu is to model a function, $f(\mathbf{X})$, which predicts the binary occupancy value for any given 3D position in continuous camera space $\mathbf{X}= (\mathbf{X}_{x}, \mathbf{X}_{y}, \mathbf{X}_{z}) \in \mathbb{R}^3$:
\begin{equation}
    f(\mathbf{X}, \mathbf{I}) = 
    \begin{cases}
            1 & \text{if $\mathbf{X}$ is inside mesh surface} \\
            0 & \text{otherwise},
        \end{cases}
    \label{eq:pifu_function_1}
\end{equation}
where $\mathbf{I}$ is a single RGB image. Since no explicit 3D volume is stored in memory during training, this approach is memory efficient, and more importantly, no discretization is needed for the target 3D volume, which is important in obtaining high-fidelity 3D geometry for the target human subjects. PIFu~\cite{saito2019pifu} models the function $f$ via a neural network architecture that is trained in an end-to-end manner. Specifically, the function $f$ first extracts a image feature embedding from the projected 2D location at $\pi(\mathbf{X}) = \mathbf{x} \in \mathbb{R}^2$, which we denote by $\Phi\left( \mathbf{x}, \mathbf{I} \right)$. Orthogonal projection is used for $\pi$, and thus $\mathbf{x} = \pi(\mathbf{X}) = (\mathbf{X}_x, \mathbf{X}_y)$. Then, it estimates the occupancy of the query 3D point $\mathbf{X}$, and thus:
\begin{equation}
f(\mathbf{X}, \mathbf{I}) = g\left( \Phi\left( \mathbf{x}, \mathbf{I} \right), Z \right),
\end{equation}
where $Z = \mathbf{X}_z$ is the depth along the ray defined by the 2D projection $\mathbf{x}$. Note that all 3D points along the same ray have exactly the same image features $\Phi\left( \mathbf{x}, I \right)$ from the same projected location $\mathbf{x}$, and thus the function $g$ should focus on the varying input depth $Z$ to disambiguate the occupancy of 3D points along the ray. In~\cite{saito2019pifu}, a Convolutional Neural Network (CNN) architecture is used for the 2D feature embedding function $\Phi$ and a Multilayer Perceptron (MLP) for the function $g$.

A large scale dataset~\cite{renderpeople} synthetically generated by rendering hundreds of high quality scanned 3D human mesh models is used to train the function $f$ in an end-to-end fashion. Unlike voxel-based methods, PIFu does not produce a discretized volume as output, so training can be performed by sampling 3D points and computing the occupancy loss at the sampled locations, without generating 3D meshes. During inference, 3D space is uniformly sampled to infer the occupancy and the final iso-surface is extracted with a threshold of 0.5 using marching cubes~\cite{lorensen1987marching}.

\paragraph{Limitations:} The input size as well as the image feature resolution of PIFu and other existing work are limited to at most 512$\times$512 and 128 $\times$ 128 in resolution respectively, due to memory limitations in existing graphics hardware. Importantly, the network should be designed such that its receptive field covers the entire image so that it can employ holistic reasoning for consistent depth inference---thus, a repeated bottom-up and top-down architecture with intermediate supervision \cite{newell2016stacked} plays an important role to achieve robust 3D reconstruction with generalization ability. This prevents the method from taking higher resolution images as input and keeping the resolution in the feature embeddings, even though this would potentially allow the network to leverage cues about detail present only at those higher resolutions. We found that while in theory the continuous representation of PIFu can represent 3D geometry at an arbitrary resolution, the expressiveness of the representation is bounded by the feature resolution in practice. Thus, we need an effective way of balancing robustness stemming from long-range holistic reasoning and expressiveness by higher feature embedding resolutions.

\subsection{Multi-Level Pixel-Aligned Implicit Function}
\label{sec:mlpifu}
We present a multi-level approach towards higher fidelity 3D human digitization that takes 1024$\times$1024 resolution images as input. Our method is composed of two levels of PIFu modules: (1) a \emph{coarse level} similar to PIFu~\cite{saito2019pifu}, focusing on integrating global geometric information by taking the downsampled $512\times512$ image as input, and producing backbone image features of $128\times128$ resolution, and (2) a \emph{fine level} that focuses on adding more subtle details by taking the original 1024$\times$1024 resolution image as input, and producing backbone image features of 512$\times$512 resolution (four times higher resolution than the implementation of \cite{saito2019pifu}). Notably, the fine level module takes 3D embedding features extracted from the coarse level instead of the absolute depth value. Our coarse level module is defined similar to PIFu, but as a modification (Sect~\ref{sec:f2b}) it also takes \emph{predicted} frontside and backside normal maps:
\begin{equation}
    f^L(\mathbf{X}) = g^L\left(\Phi^L \left( \mathbf{x}_L, \mathbf{I}_L,  \mathbf{F}_L, \mathbf{B}_L, \right), Z\right),
    \label{eq:pifu_function_2}
\end{equation}
where $\mathbf{I}_L$ is the lower resolution input and $\mathbf{F}_L$  and $\mathbf{B}_L$ are predicted normal maps at the same resolution. $\mathbf{x}_L \in \mathbb{R}^2$ is the projected 2d location of $\mathbf{X}$ in the image space of $\mathbf{I}_L$. The fine level is denoted as
\begin{equation}
    f^H(\mathbf{X}) = g^H\left(\Phi^H \left( \mathbf{x}_H, \mathbf{I}_H,  \mathbf{F}_H, \mathbf{B}_H, \right), \Omega(\mathbf{X}) \right),
    \label{eq:pifu_function_3}
\end{equation}
where $\mathbf{I}_H$, $\mathbf{F}_H$, $\mathbf{B}_H$ are the input image, frontal normal map, and backside normal map respectively at a resolution of 1024$\times$1024. $\mathbf{x}_H \in \mathbb{R}^2$ is the 2d projection location at high resolution, and thus in our case $\mathbf{x}_H = 2\mathbf{x}_L$. The function $\Phi^H$ encodes the image features from the high-resolution input and has structure similar to the low-resolution feature extractor $\Phi^L$. A key difference is that the receptive field of $\Phi^H$ does not cover the entire image, but owing to its fully convolutional architecture, a network can be trained with a random sliding window and infer at the original image resolution (i.e., 1024 $\times$ 1024). Finally, $\Omega(\mathbf{X})$ is a 3D embedding extracted from the coarse level network, where we take the output features from an intermediate layer of $g^L$.

Because the fine level takes these features from the first pixel-aligned MLP as a 3d embedding, the global reconstruction quality should not be degraded, and should improve if the network design can properly leverage the increased image resolution and network capacity. Additionally, the fine network doesn't need to handle normalization (i.e., producing a globally consistent 3D depth) and therefore doesn't need to see the entire image, allowing us to train it with image crops. This is important to allow high-resolution image inputs without being limited by memory.

\subsection{Front-to-Back Inference}
\label{sec:f2b}
Predicting the accurate geometry of the back of people is an ill-posed problem because it is not directly observed in the images. The backside must therefore be inferred entirely by the MLP prediction network and, due to the ambiguous and multimodal nature of this problem, the 3D reconstruction tends to be smooth and featureless. This is due in part to the occupancy loss (Sect.~\ref{sec:loss}) favoring average reconstructions under uncertainty, but also because the final MLP layers need to learn a complex prediction function.

We found that if we instead shift part of this inference problem into the feature extraction stage, the network can produce sharper reconstructed geometry. To do this, we predict normal maps as a proxy for 3D geometry in image space, and provide these normal maps as features to the pixel-aligned predictors. The 3D reconstruction is then guided by these maps to infer a particular 3D geometry, making it easier for the MLPs to produce details. We predict the backside and frontal normals in image space using a pix2pixHD~\cite{wang2018pix2pixHD} network, mapping from RGB color to normal maps. Similarly to recent approaches~\cite{natsume2019siclope, gabeur:hal-02242795, Smith_2019_ICCV}, we find that this produces plausible outputs for the unseen backside for sufficiently constrained problem domains, such as clothed humans.

\subsection{Loss Functions and Surface Sampling}
\label{sec:loss}
The specifics of the loss functions used can have a strong effect on the details recovered by the final model. Rather than use an average L1 or L2 loss as in~\cite{saito2019pifu}, we use an extended Binary Cross Entropy (BCE) loss~\cite{Zheng_2019_ICCV} at a set of sampled points,
\label{sec:losses}
\begin{equation}
\begin{split}
    \mathcal{L}_o = &\sum_{\mathbf{X}\in\mathcal{S}} \lambda f^*(\mathbf{X}) \operatorname{log} f^{\{L,H\}}(\mathbf{X}) \\ + &(1-\lambda) \left(1- f^*(\mathbf{X})\right) \operatorname{log} \left(1- f^{\{L,H\}}(\mathbf{X})\right),
    \label{eq:loss1}
\end{split}
\end{equation}
where $\mathcal{S}$ denotes the set of samples at which the loss is evaluated, $\lambda$ is the ratio of points outside surface in $\mathcal{S}$, $f^*(\cdot)$ denotes the ground truth occupancy at that location, and $f^{\{L,H\}}(\cdot)$ are each of the pixel-aligned implicit functions of Sect.~\ref{sec:mlpifu}. As in~\cite{saito2019pifu}, we sample points using a mixture of uniform volume samples and importance sampling around the surface using Gaussian perturbation around uniformly sampled surface points. We found that this sampling scheme produces sharper results than sampling points proportionally to the inverse of distance from the surface. In fact, a mixture of Gaussian balls on the surface has higher sampling density near regions with high curvature (up to the inverse of Gaussian ball radius). Since curvature is the second-order derivative of surface geometry, importance sampling based on curvature significantly enhances details and fidelity. 
\section{Experimental Results}
\label{sec:result}

\paragraph{Datasets.}
To obtain high-fidelity 3D geometry and corresponding images, we use RenderPeople data~\cite{saito2019pifu}, which consists of commercially available 500 high-resolution photogrammetry scans. We split the dataset into a training set of 450 subjects and a test set of 50 subjects and render the meshes with precomputed radiance transfer \cite{sloan2002precomputed} using 163 second-order spherical harmonics from HDRI Haven\footnote{\url{https://hdrihaven.com/}}. Each subject is rendered from every other degree in yaw axis with an elevation fixed with $0\degree$. Unlike \cite{saito2019pifu}, where clean segmentation mask is required, we augment random background images using COCO \cite{lin2014coco} dataset, removing the need of segmentation as pre-process. 

\paragraph{Implementation Details.}
The image encoders for both the low-resolution and high-resolution levels use a stacked hourglass network \cite{newell2016stacked} with $4$ and $1$ stacks respectively, using the modification suggested by \cite{jackson2018human} and batch normalization replaced with group normalization \cite{wu2018group}. Note that the fine image encoder removes one downsampling operation to achieve large feature embedding resolution. The feature dimensions are $128\times 128 \times 256$ in the coarse level and $512\times 512 \times 16$ in the fine level. The MLP for the coarse-level image encoder has the number of neurons of $(257, 1024, 512, 256, 128, 1)$ with skip connections at third, fourth, fifth layers. The MLP for the fine-level image encoder has the number of neurons of $(272, 512, 256, 128, 1)$ with skip connections at second and third layers. Note that the second MLP takes the output of the fourth layer in the first MLP as 3D embedding $\Omega \in \mathbb{R}^{256}$ instead of absolute depth value together with high-resolution image features $\Phi^H \left( \mathbf{x}_H, \mathbf{I}_H,  \mathbf{F}_H, \mathbf{B}_H \right) \in \mathbb{R}^{16}$, resulting in the input channel size of $272$ in total. The coarse PIFu module is pre-trained with the input image resized to $512\times512$ and a batch size of $8$. The fine PIFu is trained with a batch size of $8$ and a random window crop of size $512\times512$. We use RMSProp with weight decay by a factor of $0.1$ every $10$ epochs. Following \cite{saito2019pifu}, we use $8000$ sampled points with the mixture of uniform sampling and importance sampling around surface with standard deviations of $5$cm and $3$cm for the coarse and fine levels respectively. 

The surface normal inference uses a network architecture proposed by~\cite{johnson2016perceptual}, consisting of $9$ residual blocks with $4$ downsampling layers. We train two networks that predict frontside and backside normals individually with the following objective functions:
\begin{equation}
    \mathcal{L}_N = \mathcal{L}_{VGG} + \lambda_{l1}\mathcal{L}_{l1},
\end{equation}
where $\mathcal{L}_{VGG}$ is the perceptual loss proposed by Johnson et al. \cite{johnson2016perceptual}, and $\mathcal{L}_{l1}$ is the $l1$ distance between the prediction and ground truth normals. The relative weight $\lambda_{l1}$ is set to $5.0$ in our experiments. We use the aforementioned $450$ RenderPeople training set to generate synthetic ground truth front and backside normals together with the corresponding input images. We use Adam optimizer with learning rate of $2.0 \times 10^{-4}$ until the convergence.

\subsection{Evaluations}
\label{sec:evaluation}

\begin{figure}[t]
\centering
\includegraphics[width=\linewidth]{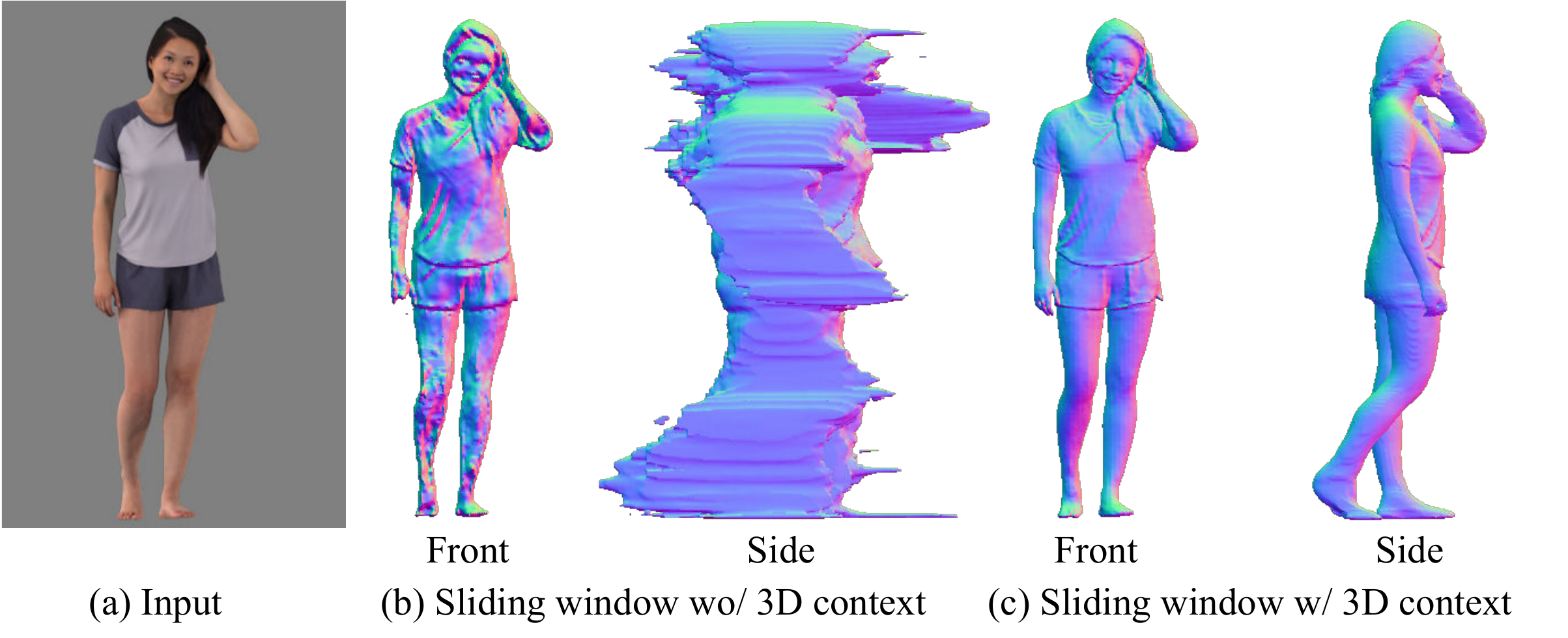}
\caption{Sliding window without 3D-aware context information, shown in (b), fails to learn plausible 3D geometry.}
\label{fig:window}
\end{figure}

\begin{figure*}[t]
\centering
\includegraphics[width=\linewidth]{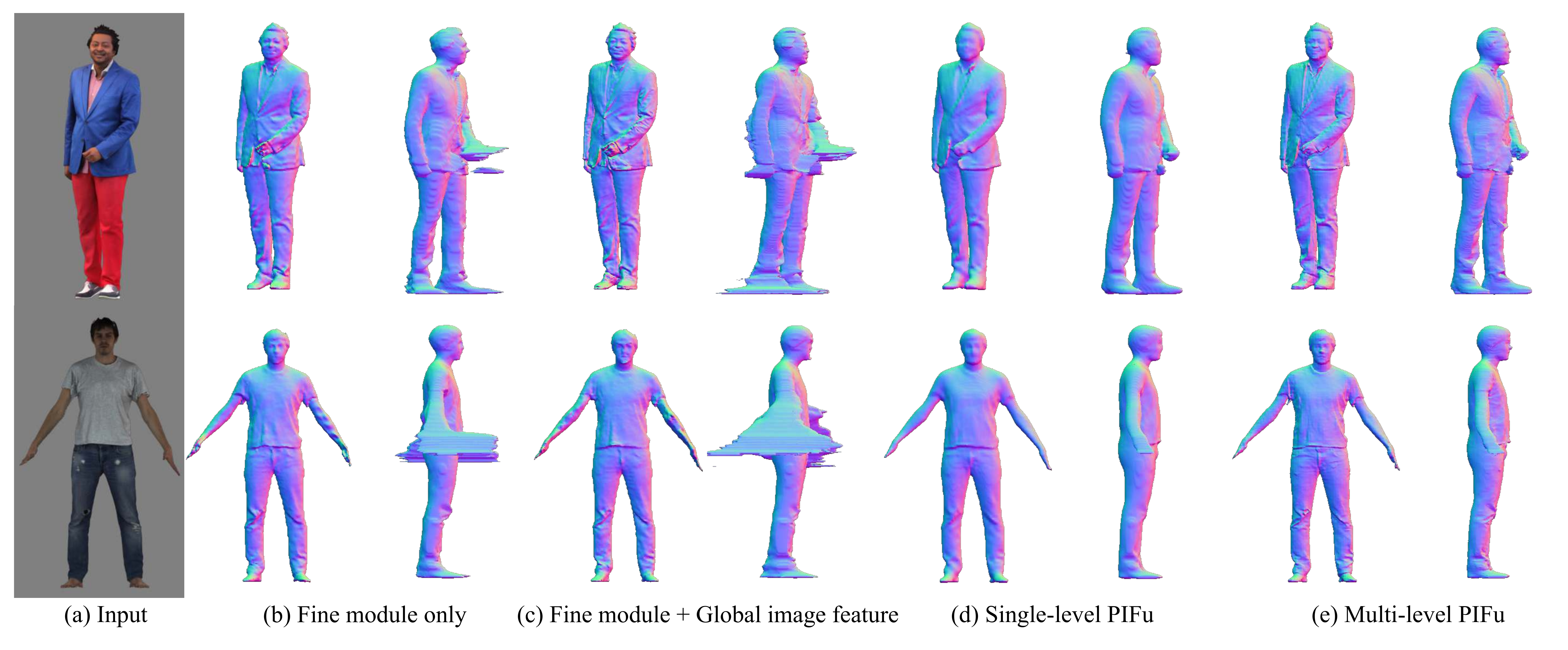}
\vspace{-15pt}
\caption{Qualitative evaluation of our multi-level pixel-aligned implicit function on samples from RenderPeople and BUFF~\cite{zhang-CVPR17} datasets. We compare the results of our method with the results of other alternative designs.}
\label{fig:ablation}
\end{figure*}

\begin{table}[t]
\resizebox{\columnwidth}{!}{
\begin{tabular}{llll|lll}
\hline
           & \multicolumn{3}{c|}{RenderPeople} & \multicolumn{3}{c}{Buff} \\
Methods    & Normal     & P2S     & Chamfer    & Normal  & P2S         & Chamfer \\ \hline
Fine module only    & 0.213      & 4.15    & 2.77        & 0.229  & 3.63         & 2.67    \\ 
Fine module + Global image feature     & 0.165      & 2.92    & 2.13       & 0.183   & 2.767    & 2.24    \\
Single PIFu    &\textbf{0.109}      & 1.45    & 1.47       & 0.134   & 1.68        & 1.76    \\
Ours (ML-PIFu, end-to-end)        & 0.117      & 1.66  & 1.55       & 0.147   & 1.88  & 1.81   \\
Ours (ML-PIFu, alternate)        &  0.111      &  \textbf{1.41}    & \textbf{1.44}       & \textbf{0.133}   & \textbf{1.63}        & \textbf{1.73}   \\
\hline
Ours with normals     & 0.107      & 1.37  & 1.43   & 0.134   & 1.63    & 1.75   \\

\end{tabular}
}
\caption{Quantitative evaluation on RenderPeople and BUFF datasets for single-view reconstruction. Units for point-to-surface and Chamfer distance are in cm.}
\label{tab:mspifu}
\end{table}

\begin{figure}[t]
\centering
\includegraphics[width=\linewidth]{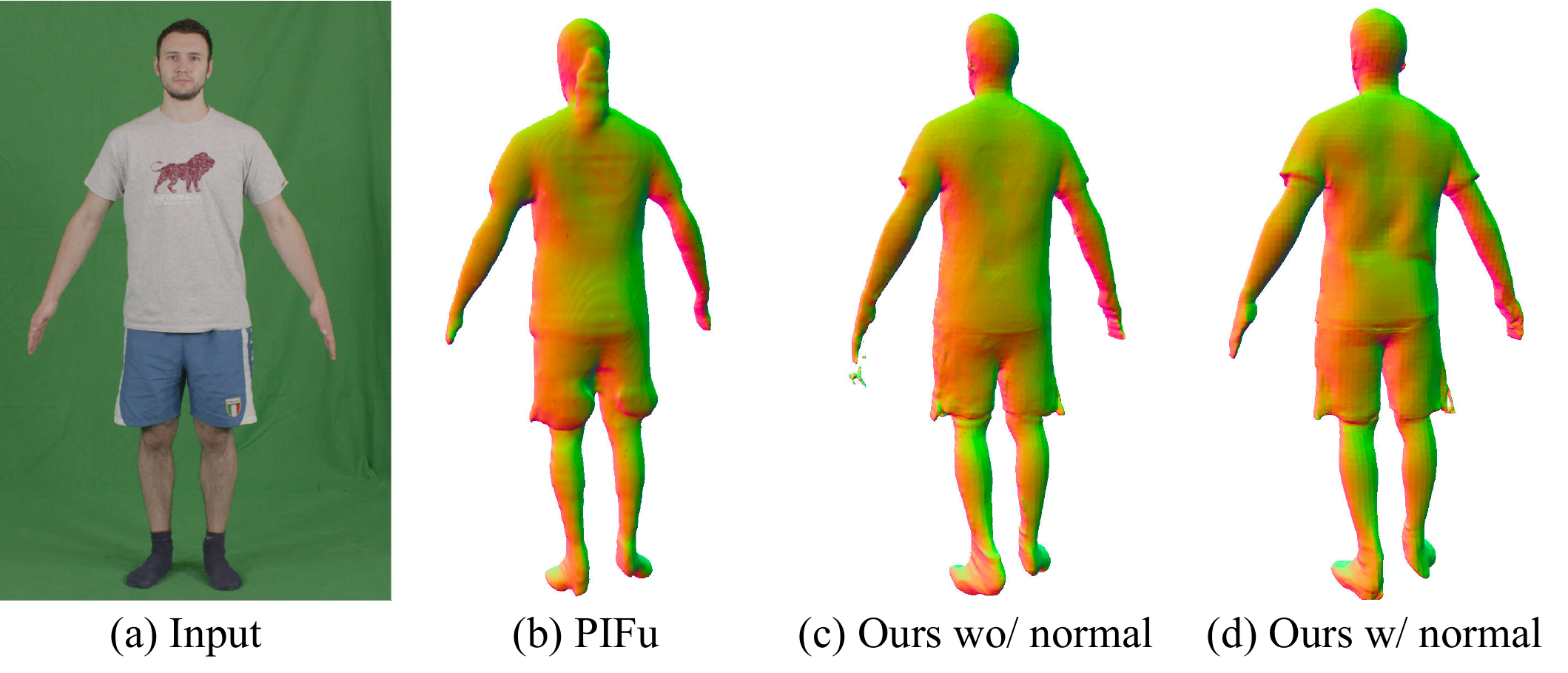}
\caption{Conditioning 3D inference with the predicted backside surface normal improves fidelity of the missing region.}
\label{fig:eval_f2bnml}
\vspace{-7pt}
\end{figure}

\paragraph{Ablation Study.}
We evaluate our multi-level pixel aligned implicit functions with several alternatives to assess the factors that contribute to achieving high-fidelity reconstructions. First, we assess the importance of 3D embedding that accounts for holistic context for high-resolution inference. To support larger input resolution at inference time, we train the network with random cropping of $512 \times 512$ from $1024 \times 1024$ images, similar to 2D computer vision tasks (e.g., semantic segmentation). We found that if our fine level module is conditioned on the absolute depth value instead of the learned 3D embedding, training with sliding window significantly degrades both training and test accuracy (see Fig. \ref{fig:window}). This illustrates that 3D reconstruction using high-resolution features without holistic reasoning severely suffers from depth ambiguity and is unable to generalize with input size discrepancy between training and inference. Thus, the combination of holistic reasoning and high-resolution images features is essential for high-fidelity 3D reconstruction.

Second, we evaluate our design choice from both robustness and fidelity perspective. To achieve high-resolution reconstruction, it is important to keep feature resolution large enough while maintaining the ability to reason holistically. In this experiment, we implement 1) a pixel-aligned implicit function using only our fine-level image encoder by processing the full resolution as input during training, 2) conditioning 1) with jointly learned global feature using ResNet34 \cite{he2016deep} as a global feature encoder in spirit to~\cite{iizuka2017globally}, 3) a single PIFu (i.e., our coarse-level image encoder) by resizing input to $512 \times 512$, 4) our proposed multi-level PIFu (two levels) by training all networks jointly (ML-PIFu, end-to-end), and 5) ours with alternate training of the coarse and fine modules (ML-PIFu, alternate). 

\begin{figure*}[t]
\centering
\includegraphics[width=\linewidth]{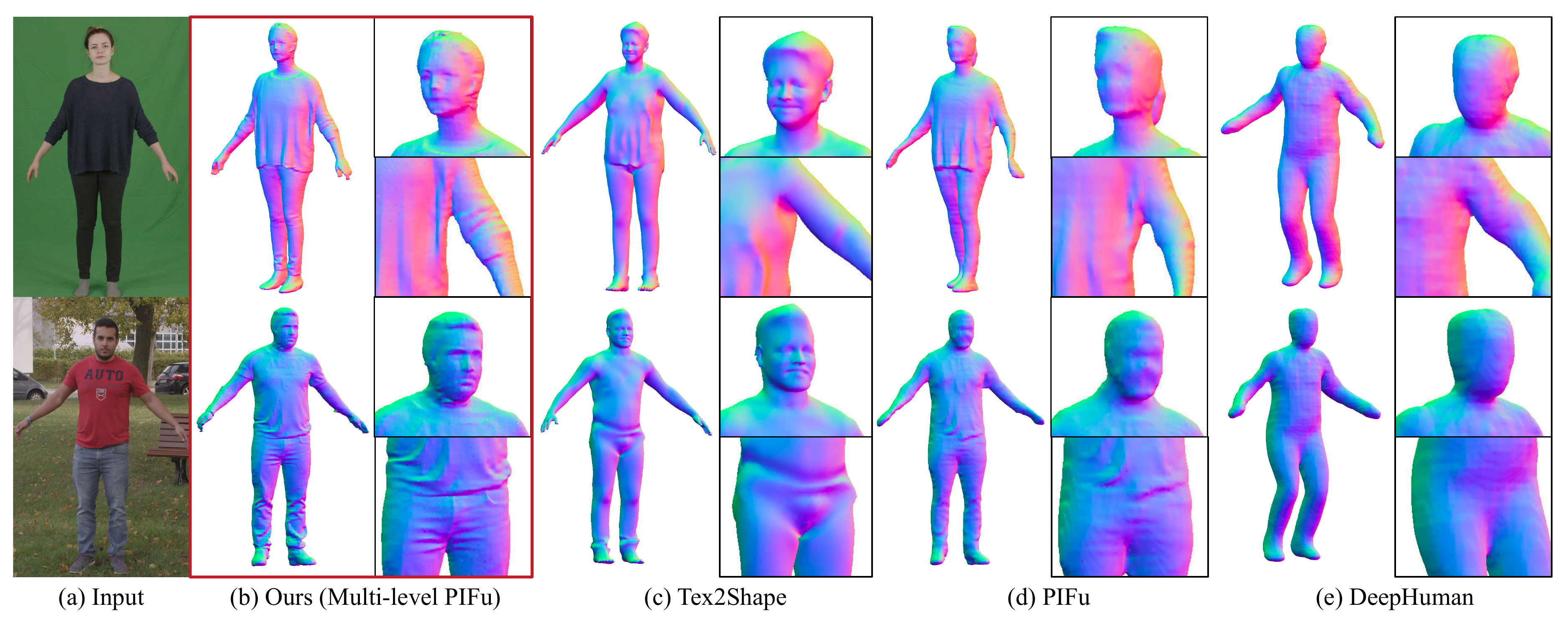}
\vspace{-14pt} 
\caption{We qualitatively compare our method with state-of-the-art methods, including (c) Tex2shape~\cite{Alldieck_2019_ICCV}, (d) PIFu~\cite{saito2019pifu}, and (e) DeepHuman~\cite{Zheng_2019_ICCV}, on the People Snapshot dataset~\cite{alldieck2018video}. By fully leveraging high-resolution image inputs, (b) our method can reconstruct higher resolution geometry compared to the existing methods.
}
\label{fig:comparison}
\vspace{-10pt}
\end{figure*}

Figure \ref{fig:ablation} and Table \ref{tab:mspifu} show our qualitative and quantitative evaluation using RenderPeople and BUFF \cite{zhang-CVPR17} dataset. We compute point-to-surface distance, Chamfer distance, and surface normal consistency using ground truth geometry. Large spatial resolution of feature embeddings ($512 \times 512$) greatly enhance local details compared with a single-level PIFu whose backbone feature resolution ($128 \times 128$) is spatially $4$ times smaller. On the other hand, due to the limited design choices for high resolution input, using local feature suffers from overfitting and robustness and generalization becomes challenging. While adding global context helps a network reason more precise geometry, resulting in sharper reconstruction, lack of precise spatial information in the global feature deteriorates the robustness. This problem becomes more critical in case of non-rigid articulated objects \cite{saito2019pifu}. Also, we found that alternatively training the coarse and fine modules results in higher accuracy than jointly training them in an end-to-end manner.  

We also evaluate the importance of inferring backside normal to recover detail on the occluded region. Figure \ref{fig:eval_f2bnml} shows that PIFu that takes only input image suffers from blurred reconstruction on the missing regions due to ambiguity. On the other hand, directly providing guidance by facilitating image-to-image translation networks significantly improves the reconstruction accuracy on both front and back side with more realistic wrinkles. Since the pixel-aligned implicit functions are computationally expensive to differentiablly render on a image plane, solving sub-problems in the image domain is a practical solution to address the completion task with plausible details.

\begin{figure*}[tb]
\centering
\includegraphics[width=\linewidth]{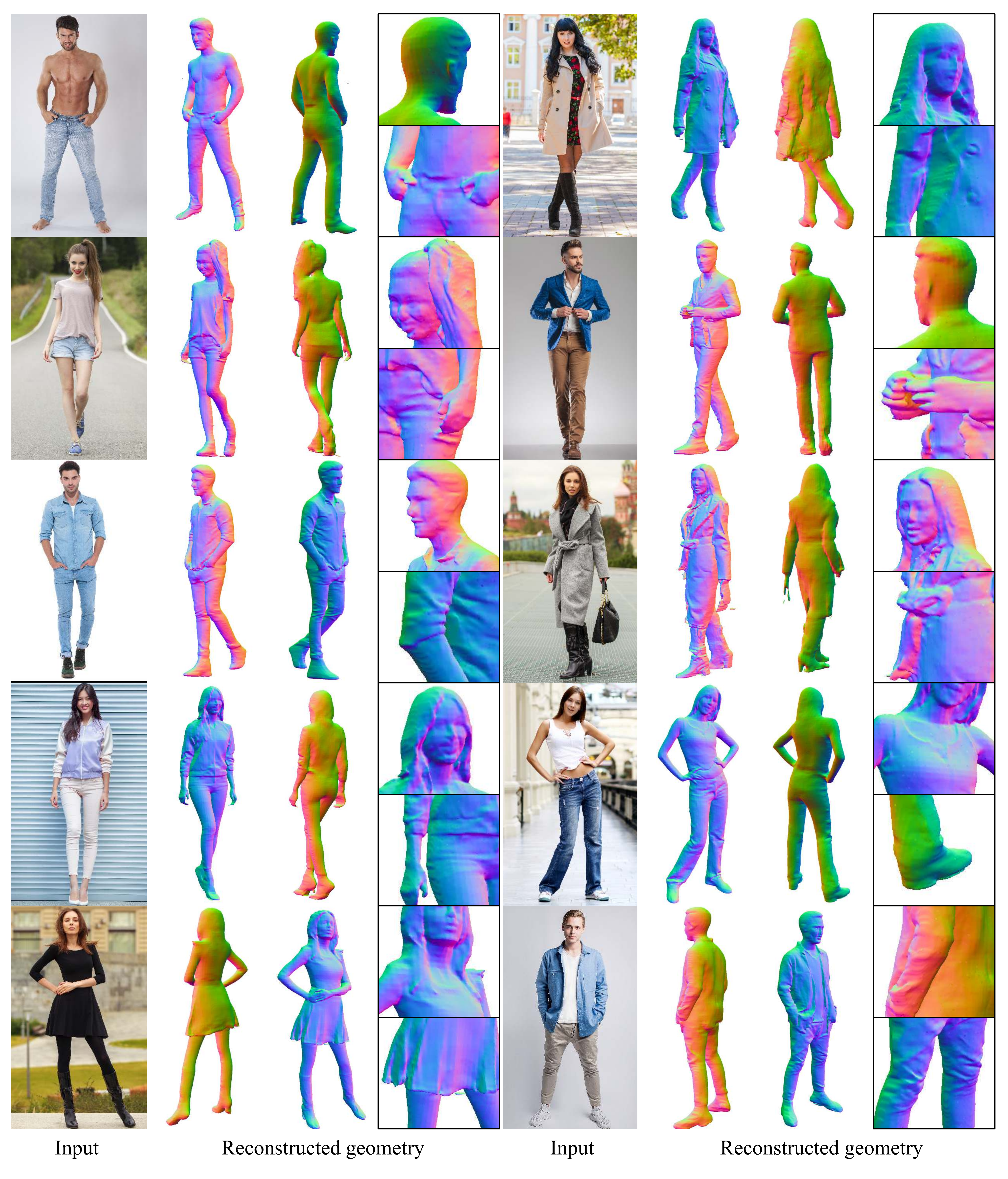}
\vspace{-5pt}
\caption{Qualitative results on Internet photos. These results demonstrate that our model trained by synthetically generated data can successfully reconstruct high-fidelity 3D from the humans in real world data.}
\label{fig:results}
\end{figure*}

\subsection{Comparisons}
\label{sec:comparisons}

We qualitatively compare our method with state-of-the-art 3D human reconstruction methods with various shape representations on the publicly available People Snapshot dataset~\cite{alldieck2018video}. The shape representations include multi-scale voxel (DeepHuman)~\cite{Zheng_2019_ICCV}, pixel-aligned implicit function (PIFu)~\cite{saito2019pifu}, and a human parametric model with texture mapping using displacements and surface normals~(Tex2shape) \cite{Alldieck_2019_ICCV}. While Tex2shape and DeepHuman adopt a coarse-to-fine strategy, the results show that the effect of refinement is marginal due to the limited representation power of the base shapes (See Fig.~\ref{fig:comparison}). More specifically, a voxel representation limits spatial resolution, and a template-based approach has difficulty handling varying topology and large deformations. Although the template-based approach \cite{Alldieck_2019_ICCV} retains some distinctive shapes such as wrinkles, the resulting shapes lose the fidelity of the input subject due to the imperfect mapping from image space to the texture parameterization using the off-the-shelf human dense correspondences map \cite{alp2018densepose}. In contrast, our method fully leverages the expressive shape representation for both base and refined shapes and directly predicts 3D geometry at a pixel-level, retaining all the details that are present in the input image. More qualitative results can be found in Figure \ref{fig:results}.

\section{Discussion and Future Work}
\label{sec:conclusion}

We present a multi-level framework that performs joint reasoning over holistic information and local details to arrive at high-resolution 3D reconstructions of clothed humans from a single image without any additional post processing or side information. Our multi-level Pixel-Aligned Implicit Function achieves this by incrementally propagating global context through a scale pyramid as an implicit 3D embedding. This avoids making premature decisions about explicit geometry that has limited prior approaches. Our experiments demonstrate that it is important to incorporate such 3D-aware context for accurate and precise reconstructions. Furthermore, we show that circumventing ambiguity in the image-domain greatly increases the consistency of 3D reconstruction detail in occluded regions. 

Since the multi-level approach relies on the success of previous stages in extracting 3D embeddings, improving the robustness of our baseline model is expected to directly merit our overall reconstruction accuracy. Future work may include incorporating human specific priors (e.g., semantic segmentation, pose, and parametric 3D face models) and adding 2D supervision of implicit surface \cite{sitzmann2019scene, liu2019learning} to further support in-the-wild inputs.

{\small
	\bibliographystyle{ieee}
	\bibliography{ms}
}

\end{document}